\documentclass[runningheads]{llncs}
\usepackage[T1]{fontenc}
\usepackage{amsmath}
\usepackage{booktabs}

\usepackage{graphicx}
\usepackage{tikz}
\def\checkmark{\tikz\fill[scale=0.4](0,.35) -- (.25,0) -- (1,.7) -- (.25,.15) -- cycle;} 

\begin{document}
\title{AVIS Engine: High Performance Networking Layer for Simulation Applications}
\titlerunning{High Performance Networking Layer for Simulation Applications}
%
\author{Amir Mohammad Zarif Shahsavan Nejad\inst{1}\orcidID{0000-0003-2969-1573} \and
Amir Mahdi Zarif Shahsavan Nejad\inst{1} \and Amirali Setayeshi\inst{2}\orcidID{0009-0000-2894-7640} \and Soroush Sadeghnejad\inst{1}\orcidID{0000-0001-8286-7756} }
\authorrunning{A. Zarif, et al.}
%
\institute {
Bio-Inspired System Design Lab, Department of Biomedical Engineering, Amirkabir University of Technology (Tehran Polytechnic), Tehran 15914, Iran \\ \and
Department of Electrical Engineering, Amirkabir University of Technology (Tehran Polytechnic), Tehran 15914, Iran \\
\textbf{correspondingauthor}: \email{s.sadeghnejad@aut.ac.ir}
}

\maketitle              
\begin{abstract}
Autonomous vehicles are one of the most popular and also fast-growing technologies in the world. As we go further, there are still a lot of challenges that are unsolved and may cause problems in the future when it comes to testing in real world. Simulations on the other hand have always had a huge impact in the fields of science, technology, physics, etc. The simulation also powers real-world Autonomous Vehicles nowadays. Therefore, We have built an Autonomous Vehicle Simulation Software - called AVIS Engine - that provides tools and features that help develop autonomous vehicles in various environments. AVIS Engine features an advanced input and output system for the vehicle and includes a traffic system and vehicle sensor system which can be communicated using the fast networking system and ROS Bridge.

\end{abstract}
\section{Introduction}
Artificial Intelligence (AI) is one of the biggest topics in the world today and plays an important role in people's lives. Hence, it should be more and more intelligent and safe to be able play its role perfectly. Simulations have always been a part of robotics and AI such as industrial robots, surgical and medical robots \cite{sadeghnejad2019validation}, humanoid robots, etc. which can become extremely complex and difficult as we progress \cite{burkhard2002road} so is the RoboCup goal for 2050.
RoboCup also involved simulators for wheeled robot since the beginnings \cite{gerndt2015humanoid,kitano1997robocup} and also made the simulation to more robotics research groups till this day \cite{gerndt2015humanoid}.
Autonomous vehicles (AVs) are one of the applications of AI in the real world and are designed to help people travel safer, faster, easier, and cheaper than non-electric and non-autonomous vehicles.
AVs also promise to be more environmentally friendly, as they are electrically powered and produce fewer pollutants than internal combustion engine vehicles \cite{Nimalsiri}.

In most cases, many experiments are required to approximately guarantee the safety of these modern AVs. The experiments can also be expensive. Testing and experimenting in simulation environments is a solution to build a much more secure AV that costs less compared to building it in the real world.

Simulating different objects and machines can be very beneficial because it is easy to study them. On the other hand, real-world research can be very expensive. For example, in the field of AVs, experimenting with a real car is much more expensive and time consuming than a simulation given the cost of the tools and equipment needed to build one AV.
Simulations can be beneficial in studying algorithms, training machine learning models, finding different solutions and approaches to solving a problem, and identifying "driving" variables. \cite{intsim}

For instance, obstacle avoidance or path planning in the real physical form of the vehicle can be extremely difficult and also expensive, and certainly errors can have some unavoidable consequences. Usually, testing algorithms in simulator can be very time-saving and also cheaper than other solutions. Resetting and quickly testing other conditions without cost is one of the other advantages of using simulators, which can be useful in machine learning and deep learning solutions.

All different scenarios and conditions should be considered when a technology is based on AI, and one solution to this is the use of simulations. Some researchers use videos to develop lane recognition keeping algorithms, and there is no other input or output from the vehicle. But a simulator can actually process those inputs and outputs. Nowadays, the trend of AI is growing so fast and there are so many things that are related to AI in some way and have been extended to an application of AI.

There are other simulators for autonomous vehicles that also aim to support and advance research in this area, including Apollo \cite{Apollo}, AutonoViSim \cite{avs}, CARLA  \cite{Dosovitskiy17}, Force-based, GTAV, SimMobilityST \cite{simost}, SUMMIT \cite{9197228}, SUMO \cite{sumo} and TORCS \cite{torcs}. Table \ref{table:comp} shows a comparison between the existing simulators and platforms.

\begin{table}
\centering
\caption{A Comparison between existing simulators (Derived from \cite{9197228})}
\label{table:comp}
\begin{tabular*}{\linewidth}{c|c|c|c|c} 
\toprule
Simulator          & Real-world Maps & Unregulated Behaviors & Dense Traffic & Realistic Visuals  \\ 
\midrule
Apollo             & $\times$        & $\times$              & $\times$      & \checkmark             \\
AutonoViSim        & $\times$        & \checkmark                & \checkmark        & \checkmark             \\
CARLA              & $\times$        & $\times$              & $\times$      & \checkmark             \\
Force-based        & $\times$        & $\times$              & \checkmark        & \checkmark             \\
GTAV               & $\times$        & $\times$              & $\times$      & \checkmark             \\
SimMobilityST      & \checkmark          & $\times$              & \checkmark        & $\times$           \\
SUMMIT             & \checkmark          & \checkmark                & \checkmark        & \checkmark             \\
SUMO               & $\times$        & $\times$              & \checkmark        & \checkmark             \\
AVIS Engine (Ours) & \checkmark          & \checkmark                & \checkmark        & \checkmark             \\
\bottomrule
\end{tabular*}
\end{table}

We have developed simulation software for AVs called AVIS Engine (Autonomous Vehicles Intelligent Simulation Software)\footnote{AVIS Engine Website:\\
\texttt{https://avisengine.com}} that allows researchers to explore get their hands on the research and development of AVs in a way that makes the process truly feasible and fast. CARLA and Microsoft AirSim \cite{airsim2017fsr} that are notable in the AV simulation. Both Microsoft AirSim and CARLA are powered by the Unreal Engine. Both simulators provide powerful realistic 3D simulations. AVIS Engine uses Unity 3D as its main engine. Both Unreal Engine and Unity 3D are suitable for high-fidelity rendering in applications such as robotics simulations. \cite{GuerraTMRK19}).\\
The AVIS Engine can also be used to experiment with the sound of the environment, i.e. it provides multiple listeners placed around the vehicle as microphones that pick up the sound of any object in the scene such as other vehicles, pedestrians, ambulances and other criteria.

The AVIS Engine includes most of the features required for the AI development of the AVs are helpful for researchers and developers developing AVs and AI.

\section{AVIS Engine}
\begin{figure}[!htb]
\centering
\includegraphics[width=\linewidth]{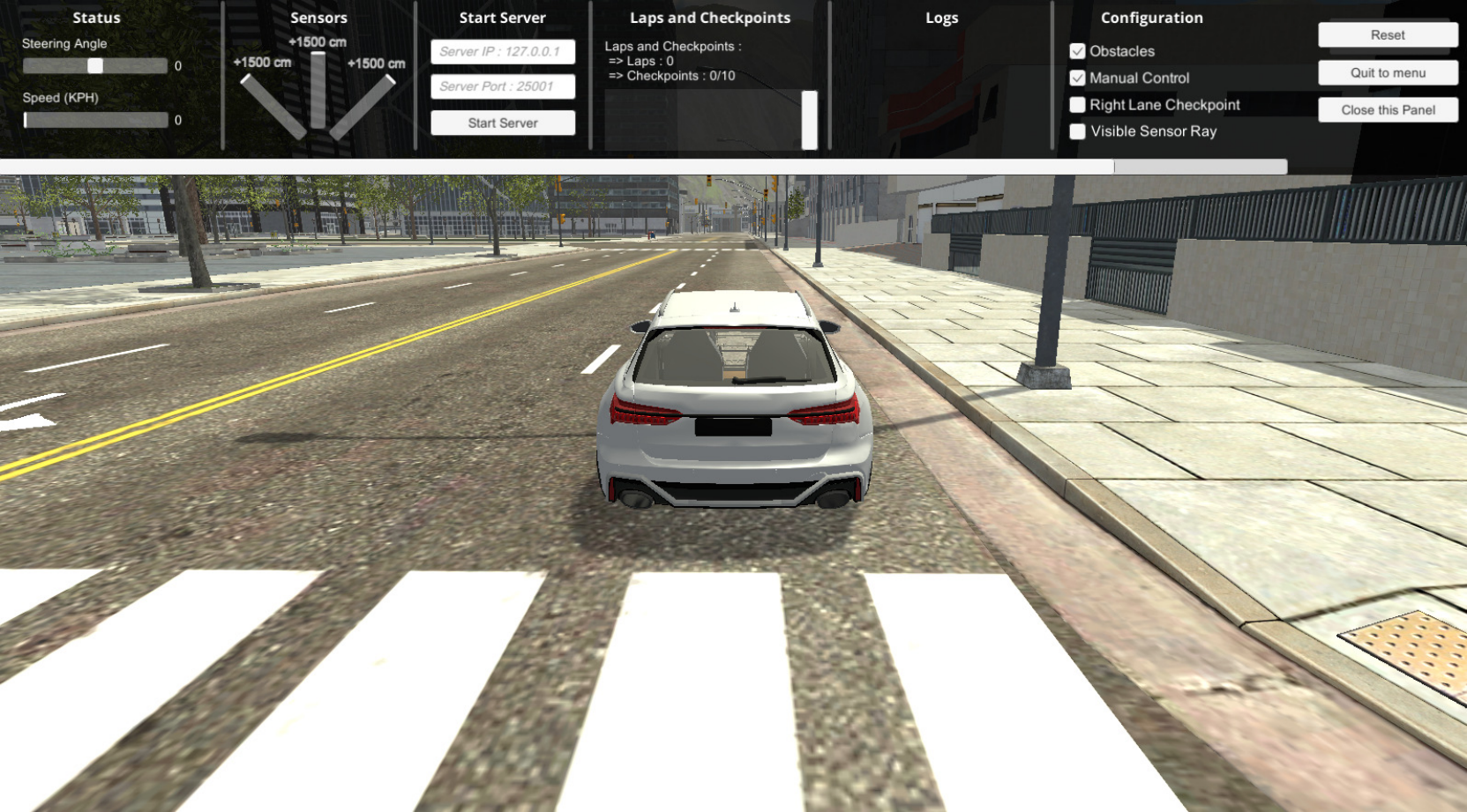}
\caption{A screenshot of the main urban environment in AVIS Engine. The panel at the top of the screen is used for configuration and visualization of the necessary data of the vehicle}
\label{fig:simulator}
\end{figure}

AVIS Engine is a robust and fast simulation software for autonomous vehicles that extremely simplifies the development of AVs and is quite easy to use. Figure \ref{fig:simulator} shows the main interface of the software.
The simulator uses its own messaging system that relies on the Transmission Control Protocol (TCP). The program transmits data and commands over a TCP connection where the simulator is the server peer of the connection and the program is the client peer of the connection \cite{bib8}.

In order to ensure reliable, practical, and secure data transfer and to allow smooth communication between the server (simulator) and the client (script), we have chosen TCP instead of UDP for the implementation of this system. In general, clients can be written in any technology or programming language that supports TCP. So far, AVIS Engine communicates with Python, C++, ROS (Robotic Operating System) and MATLAB \cite{matlab} and is flexible to support other technologies in the future.

The simulator itself is based on Unity \cite{unity} and the C\# programming language, which enabled the development of powerful cross-platform simulation software. The simulator includes and offers a wide range of features and key values, to name a few:

\subsection{High Performance}
AVIS Engine is compatible with low-end PCs and is capable of running smoothly on them. This is because the software core is lightweight and the underlying network system is fast and optimized for such scenarios.

\subsection{Fast Networking}
AVIS Engine messaging system aims to deliver and transmit data in an efficient manner, resulting in a fast and agile network. Section \ref{section:networking} explains the networking of the simulator.
 
\subsection{Global Positioning System (GPS)}
The GPS unit of AVs is one of the most important components that plays an important role in guiding and navigating the vehicle and can be used in various applications such as navigation systems and path planning. The cost of precise GPS receivers is the major drawback of GPS for these autonomous applications \cite{Campbell}.

\subsection{Semantic Camera}
\begin{figure}[h]
\centering
\includegraphics[width=\linewidth]{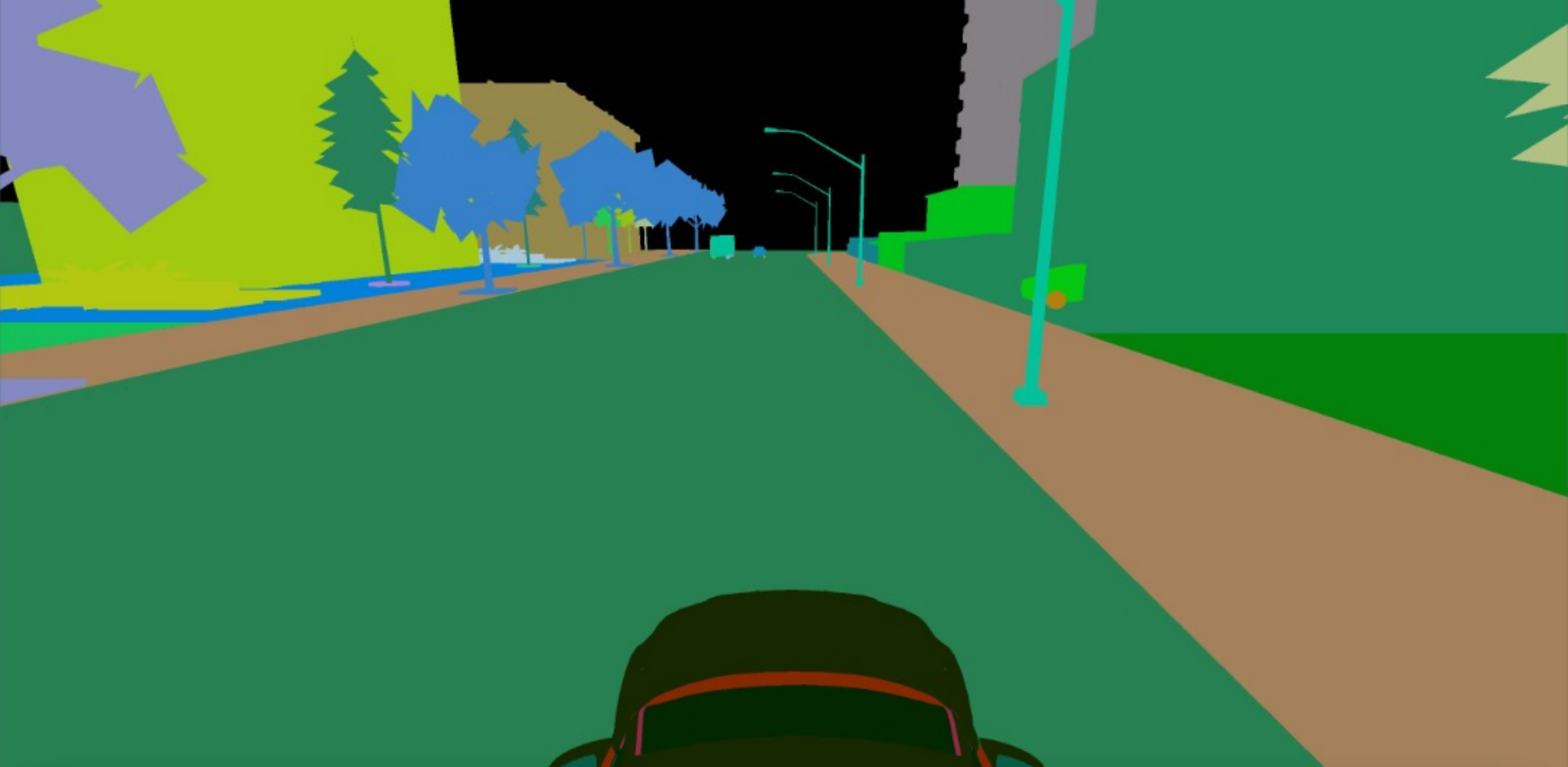}
\caption{Semantic view of the front camera}
\label{fig:semantic}
\end{figure}

A semantic segmentation that can be used for many purposes, such as problems related to machine learning and deep learning. It can also be used to improve photorealism (Figure \ref{fig:semantic}).

\subsection{Depth camera}
\begin{figure}[h]
\centering
\includegraphics[width=\linewidth]{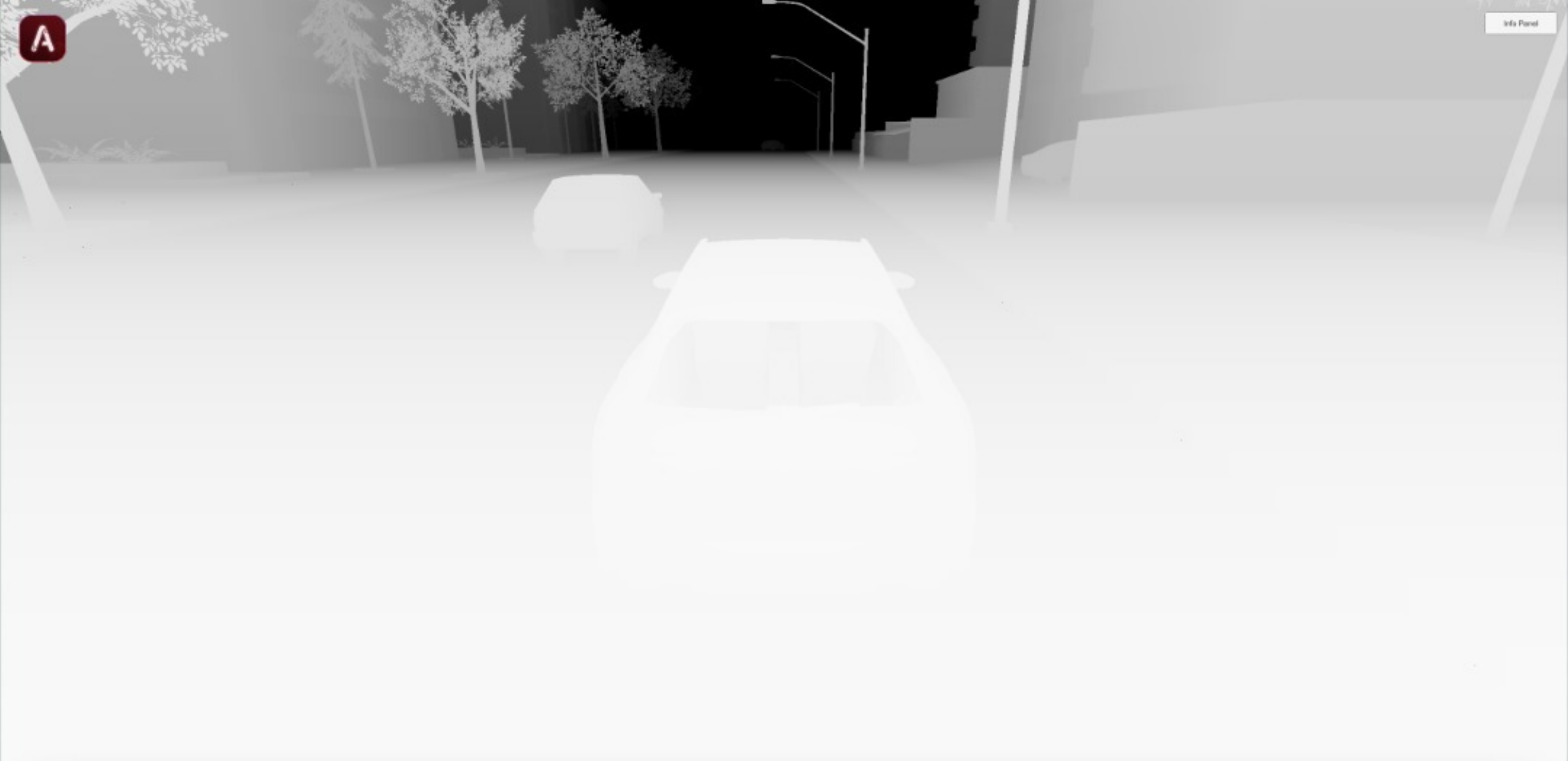}
\caption{Depth camera view in the simulator interface}
\label{fig:depthview}
\end{figure}

Concretely, there is no depth information in a normal camera, this is where the depth camera comes to the rescue. The depth camera is used for depth estimation and provides a better understanding of the environment. With this feature, a third dimension is added to the camera information besides the 2-D image, namely the depth (Figure \ref{fig:depthview}).

\subsection{Built-in filter for road features}
\begin{figure}[h]
\centering
\includegraphics[width=\linewidth]{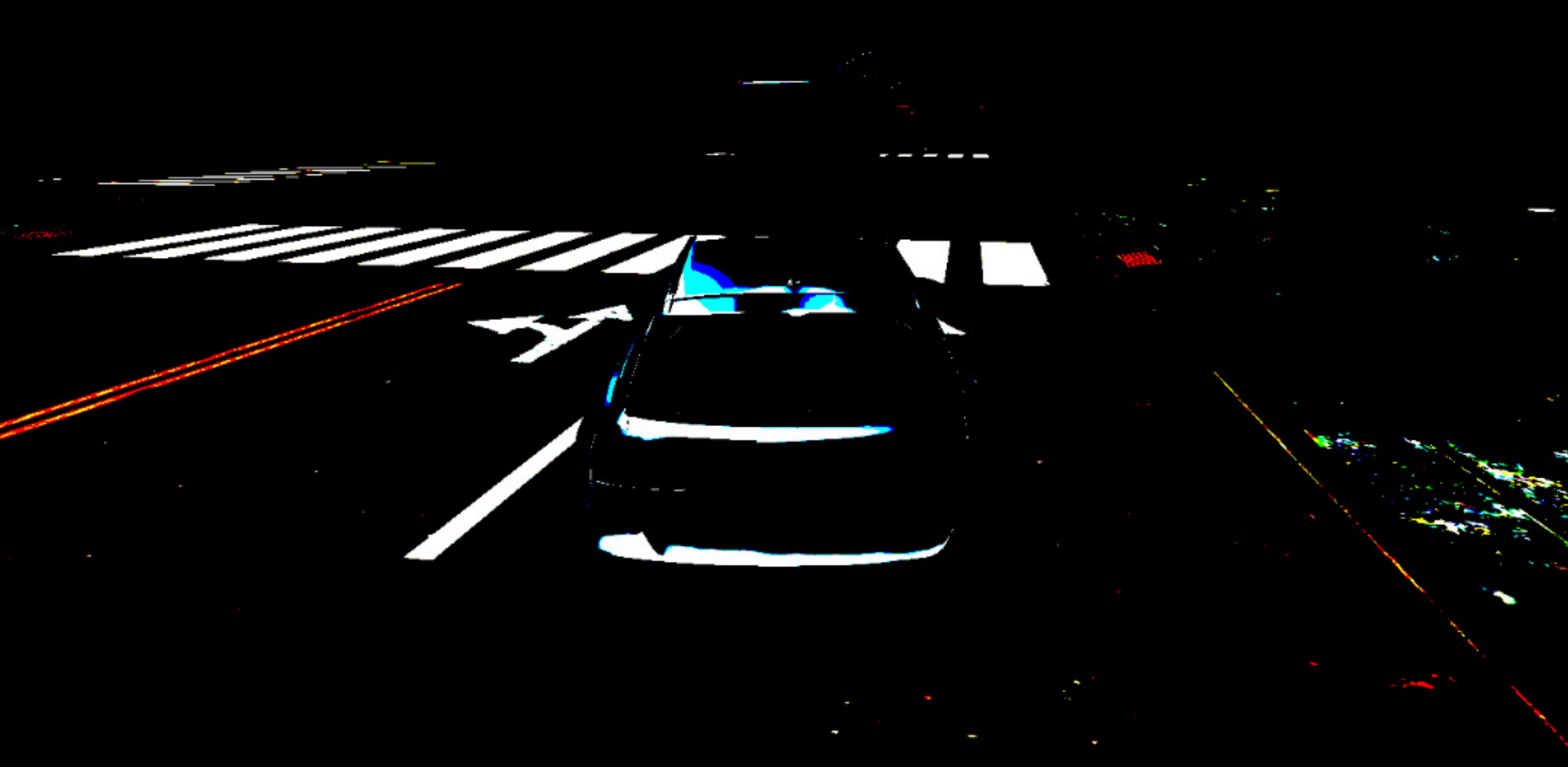}
\caption{Road feature filter view}
\label{fig:roadfeature}
\end{figure}

In addition to the various cameras and sensors, the simulator has a built-in road feature filter that highlights road features captured from the camera (Figure \ref{fig:roadfeature}).

\subsection{Depth camera}
Concretely, there is no depth information in a normal camera, this is where the depth camera comes to the rescue. The depth camera is used for depth estimation and provides a better understanding of the environment. With this feature, a third dimension is added to the camera information besides the 2-D image, namely the depth (Figure \ref{fig:depthview}).

\subsection{Built-in filter for road features}
In addition to the various cameras and sensors, the simulator has a built-in road feature filter that highlights road features captured from the camera (Figure \ref{fig:roadfeature}).

\subsection{Light Detection and Ranging (LIDAR)}

\begin{figure}[!ht]
\centering
\includegraphics[width=\linewidth]{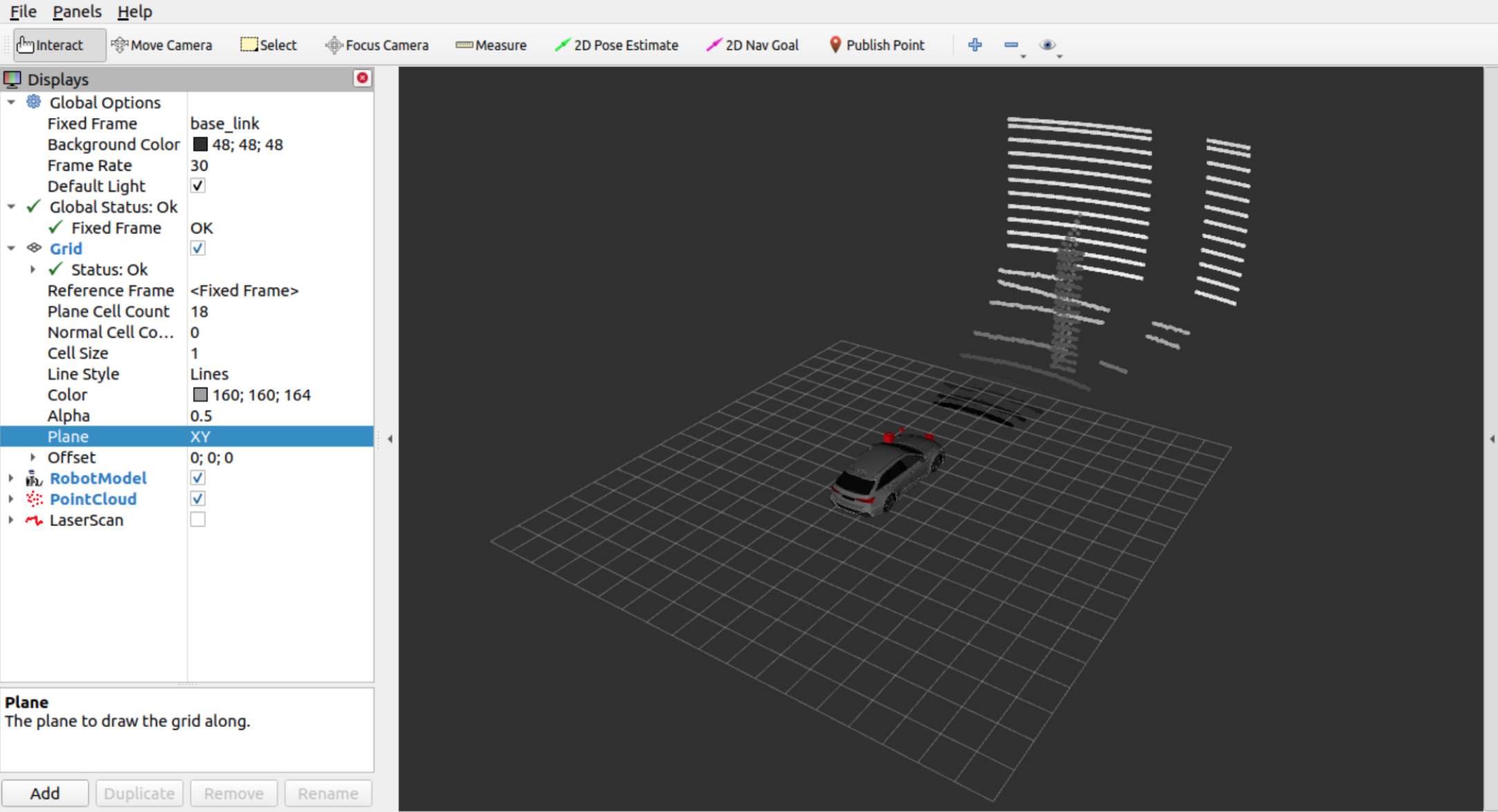}
\caption{Visualization of the LIDAR Sensor}
\label{fig:rviz}
\end{figure}

Detect and understand the environment in even greater detail. The sensor has been simulated as a set of raycasts and the output is represented as a point cloud and can be visualized in RViz \cite{Kam2015RVizAT} using the AVIS Engine ROS Bridge (Figure \ref{fig:rviz}). It is also customizable in 4 general specifications \cite{raj2020survey} including: 
\begin{enumerate}
 		\item Ranging Specification
   		\item Physical Specifications
   		\item Laser Specification
   		\item Optical Specification 
\end{enumerate}

\subsection{Traffic System}
\begin{figure}
\centering
\includegraphics[width=\linewidth]{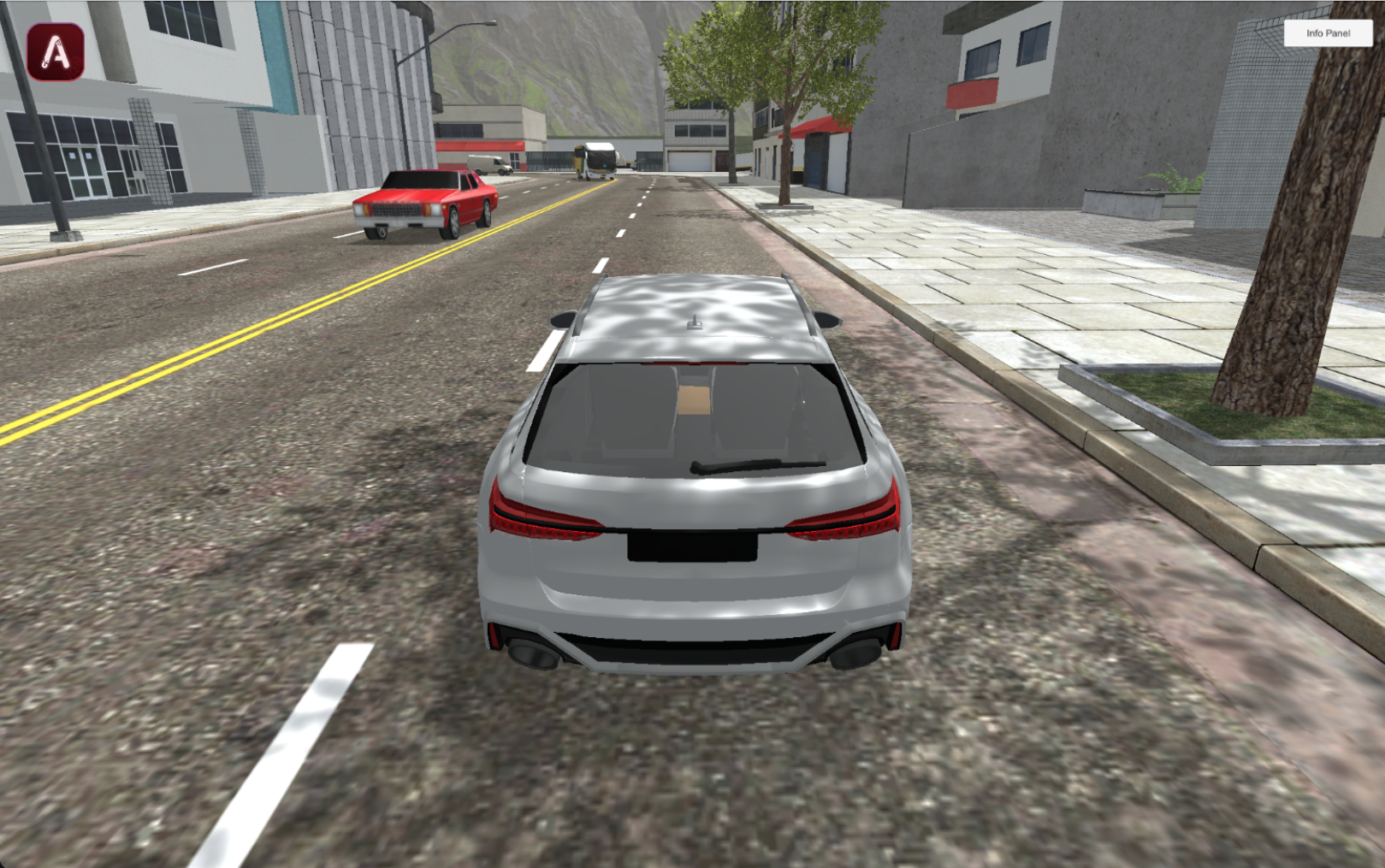}
\caption{AVIS Engine Traffic System}
\label{fig:traffic}
\end{figure}

Driving in an empty city with no other vehicles on the roads is not much of a challenge. But when it comes to driving between other cars on the road, it becomes even more difficult. AVIS Engine provides a customizable traffic system that allows a more detailed study of AV (Figure \ref{fig:traffic}).

\subsection{Vehicle communication systems}
Inter-vehicle communication is one of the most important components for AVs and smart vehicles, and hopes to cover and improve issues around vehicle safety. \cite{Luo}. The AVIS Engine has a system that provides a vehicle communication system for vehicle communication studies.

\subsection{ROS Bridge}
ROS is a powerful software development kit for robotics \cite{ROS,ROSPaper}. The AVIS Engine features a compatibility layer that connects directly to ROS and includes many built-in tools as well as community-developed tools and packages. This allows for efficient and practical visualization, development of control systems, simulations, and hardware implementations for which comple
\section{Networking}
\label{section:networking}
\begin{figure}
\centering
\includegraphics[width=\linewidth]{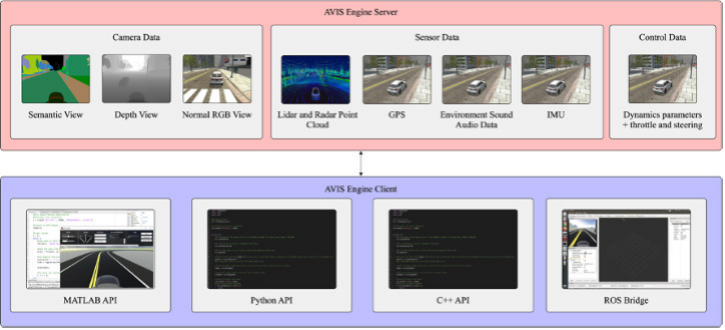}
\caption{Founding Layers and various parts of the simulator}
\label{fig:simulatorlayers}
\end{figure}

The AVIS Engine is built on two basic building blocks: 1. AVIS Engine Server software, 2. AVIS Engine Client (Figure  \ref{fig:simulatorlayers}). The server software is responsible for the physics simulations, the 3D environment, the sound, the vehicle model, the sensor simulations and all simulations related to the vehicle. The client, on the other hand, is responsible for sending the client commands to the server and receiving the response through an interface. The client can be written in any programming language or platform that can establish a TCP connection. The most important and extensive API developed for AVIS Engine is written in Python programming language. Communication between these two basic building blocks requires networking and is also the bottleneck of the whole system.

Therefore, a fast networking system is crucial to achieve fast and stable communication between the client and the server. Many issues must be considered in its development. Dealing with different approaches to improve and optimize the network such as compression algorithms.

We have experimented with different approaches to optimize the network layer of the software, since a fast networking system is crucial.
In early versions of AVIS Engine, the software used Portable Network Graphics (PNG) format to transmit images from the vehicle's camera sensors. It was also low resolution (256 pixels * 256 pixels) and slow overall. After performing techniques, the simulator is now able to transmit 1080p images to the client without any speed degradation. Here are some techniques that were used to optimize the network layer:

\subsection{Image Compression}
The first and most effective step in the optimization process was the lossy compression and conversion of the camera images to JPEG \footnote {Joint Photographic Experts Group} format before they were sent over. This technique had a great impact on increasing the speed of data transmission and reducing the delay. An example of the reduced image of the camera sensor is described in the Table \ref{table:networking}.

\subsection{Color-Space Conversion}
Color-space conversion is a widely used technique and is frequently used in computer vision and artificial intelligence (see \cite{billaut2018colorunet} for an example of an artificial intelligence study). Also used for television systems and high-definition television. This technique allowed us to slightly compress the image camera data in the simulator, but it resulted in data loss. The conversion to YCbCr is derived as follows and is defined by ITU-R BT.601 \cite{yuv,Ibraheem,yang}:
\begin{enumerate}
\item RGB to YCbCr Conversion:
\begin{equation}
\begin{bmatrix}
Y'\\ 
Cb\\ 
Cr
\end{bmatrix}
=
\begin{bmatrix}
0.299 & 0.587 & 0.114\\ 
-0.169 & -0.331 & 0.499\\ 
0.499 & -0.418 & -0.0813 
\end{bmatrix}
\begin{bmatrix}
R\\ 
G\\ 
B
\end{bmatrix}
+
\begin{bmatrix}
0\\ 
128\\ 
128
\end{bmatrix}
\end{equation}
Which can be also written as:

\begin{equation}
\begin{cases}
Y = 0.299R + 0.587G + 0.114B 
\\ 
Cb = -0.169R + -0.332G + 0.500B + 128 
\\ 
Cr = 0.500R + -0.419G + -0.0813B + 128
\end{cases}
\end{equation}
\item YCbCr to RGB Conversion:
\begin{equation}
\begin{bmatrix}
R\\ 
G\\ 
B
\end{bmatrix}
=
\begin{bmatrix}
1 & 0 & 1.402\\ 
1 & -0.334 & -0.714\\ 
1 & 1.772 & 0 
\end{bmatrix}
\begin{bmatrix}
Y\\ 
Cb - 128\\ 
Cr - 128
\end{bmatrix}
\end{equation}
Which can be written as:
\begin{equation}
\begin{cases}
R = Y + 1.407 \times (V - 128)
\\ 
G = Y - 0.345 \times (Cb - 128) - (0.716 \times (Cr - 128))
\\ 
B = Y + 1.779 \times (Cb - 128)
\end{cases}
\end{equation}
\end{enumerate}

\subsection{GZip Compression}
This method is usually effective in compressing the data. However, in the case of our simulator, it was unexpectedly ineffective and the change in data size was small.
\begin{table}
\centering
\caption{Detailed information about how much data was compressed after each technique was performed}
\label{table:networking}
\begin{tabular}{lllll} 
\toprule

Technique                             & \begin{tabular}[c]{@{}l@{}}Raw Camera Data\\ Uncompressed\\ (303459 Bytes)\end{tabular} & Change in Data Size  &  &   \\ 
\midrule
Image Compression (PNG to JPEG)       & 32932 Bytes                                                                             & $\sim$90\% Reduction &  &   \\
Color-Space Conversion (RGB to YCbCr) & 100141 Bytes                                                                            & $\sim$33\% Reduction &  &   \\
GZip Compression                      & 12138 Bytes                                                                             & $\sim$4\% Reduction  &  &   \\
\bottomrule
\end{tabular}
\end{table}

\section{A Performance Comparison}
This section compares the performance of AVIS Engine with other available AV simulation software. Note that all tests are performed on a laptop with the specifications listed in the Table \ref{table:specs} . In particular, these tests compare the AVIS engine to the CARLA simulator, a high-quality simulator for autonomous vehicles. Both simulators provide a pedestrian system and a traffic system, but to test performance, these features were not used.

\begin{table}[!htb]
\centering
\caption{Test system specifications}
\label{table:specs}
\begin{tabular}{ll} 
\hline
Configuration        & Detail                               \\\hline

CPU                  & Intel® Core™ i5-5257U CPU @ 2.70GHz  \\
GPU                  & Intel® Iris Graphics 6100 1536 MB    \\
RAM                  & 8 GB                                 \\
Screen Resolution    & 2560 x 1600                          \\
Simulator Resolution & 1920 x 1080                          \\
\hline

\end{tabular}
\end{table}

In the tests, no other cars or special objects are spawned in the scene, so there is only simulator running with the main vehicle spawned and no client connected.

Table \ref{table:performanceresults} shows the results of the multiple tasks performed with each simulator. The highest graphics setting in AVIS Engine is "Ultra" and the equivalent setting in CARLA is "Epic". The lowest setting in AVIS Engine is called "Fastest" and in CARLA it is called "Low". As the experiments show, the AVIS Engine simulator is capable of running at high FPS even on a low-end laptop, which can be useful when it comes to using efficient simulation software for work or a study. With the AVIS Engine it is possible to achieve over 110 FPS on a dedicated NVIDIA GeForce 1070 graphics card.

\begin{table}
\centering
\caption{Performance results from each simulator in different settings}
\label{table:performanceresults}
\begin{tabular}{lllll} 
\toprule
Simulator   & \multicolumn{2}{l}{FPS in Lowest Settings} & \multicolumn{2}{l}{FPS in Highest  Settings}  \\ 
\midrule
            & Avg. & Top                                 & Avg. & Top                                    \\ 
\cmidrule(lr){2-5}
AVIS Engine & 57   & 65                                  & 42   & 50                                     \\
CARLA       & 7    & 10                                  & 2    & 5                                      \\
\bottomrule
\end{tabular}
\end{table}

\section{Conclusion}
We have studied one state-of-the-art techniques that powers the development autonomous vehicles which is simulation, and its necessity.
We have also described the AVIS Engine Simulator and its capabilities, features, and capabilities, and analyzed the simulator's good performance. In addition, we have unpacked the ideas behind its fast networking, with the results clearly showing how the techniques described in the paper, such as compression algorithms and color-space conversions, have helped to reduce the data load and thus enormously optimize the performance of the software.

\section*{Acknowledgment}
This work was supported by Bio-Inspired System Design Lab (BInSDeLa) at Amirkabir University of Technology.

%
%
%
\bibliographystyle{splncs04}
\bibliography{citation}

\begin{thebibliography}{10}
\providecommand{\url}[1]{\texttt{#1}}
\providecommand{\urlprefix}{URL }
\providecommand{\doi}[1]{https://doi.org/#1}

\bibitem{matlab}
Matlab website (2022), \url{https://www.mathworks.com/products/matlab.html}

\bibitem{torcs}
Torcs - the open racing car simulator (2022),
  \url{https://sourceforge.net/projects/torcs/}

\bibitem{unity}
Unity real-time development platform (2022), \url{https://unity.com/}

\bibitem{ROS}
Why ros? (2022), \url{https://www.ros.org/blog/why-ros/}

\bibitem{Apollo}
Apollo: Apollo game engine based simulator - an autonomous vehicle simulator
  built with unity (2022), \url{https://developer.apollo.auto/gamesim.html}

\bibitem{avs}
Best, A., Narang, S., Pasqualin, L., Barber, D., Manocha, D.: Autonovi-sim:
  Autonomous vehicle simulation platform with weather, sensing, and traffic
  control. In: 2018 IEEE/CVF Conference on Computer Vision and Pattern
  Recognition Workshops (CVPRW). pp. 1161--11618 (2018).
  \doi{10.1109/CVPRW.2018.00152}

\bibitem{billaut2018colorunet}
Billaut, V., de~Rochemonteix, M., Thibault, M.: Colorunet: A convolutional
  classification approach to colorization. arXiv preprint arXiv:1811.03120
  (2018)

\bibitem{burkhard2002road}
Burkhard, H.D., Duhaut, D., Fujita, M., Lima, P., Murphy, R., Rojas, R.: The
  road to robocup 2050. IEEE Robotics \& Automation Magazine  \textbf{9}(2),
  31--38 (2002)

\bibitem{9197228}
Cai, P., Lee, Y., Luo, Y., Hsu, D.: Summit: A simulator for urban driving in
  massive mixed traffic. In: 2020 IEEE International Conference on Robotics and
  Automation (ICRA). pp. 4023--4029 (2020).
  \doi{10.1109/ICRA40945.2020.9197228}

\bibitem{Campbell}
Campbell, S., O'~Mahony, N., Krpalkova, L., Riordan, D., Walsh, J., Murphy, A.,
  Ryan, C.: Sensor technology in autonomous vehicles : A review. pp.~1--4 (06
  2018). \doi{10.1109/ISSC.2018.8585340}

\bibitem{Dosovitskiy17}
Dosovitskiy, A., Ros, G., Codevilla, F., Lopez, A., Koltun, V.: {CARLA}: {An}
  open urban driving simulator. In: Proceedings of the 1st Annual Conference on
  Robot Learning. pp. 1--16 (2017)

\bibitem{gerndt2015humanoid}
Gerndt, R., Seifert, D., Baltes, J.H., Sadeghnejad, S., Behnke, S.: Humanoid
  robots in soccer: Robots versus humans in robocup 2050. IEEE Robotics \&
  Automation Magazine  \textbf{22}(3),  147--154 (2015)

\bibitem{GuerraTMRK19}
Guerra, W., Tal, E., Murali, V., Ryou, G., Karaman, S.: Flightgoggles:
  Photorealistic sensor simulation for perception-driven robotics using
  photogrammetry and virtual reality. In: 2019 IEEE/RSJ International
  Conference on Intelligent Robots and Systems, IROS 2019, Macau, SAR, China,
  November 3-8, 2019. pp. 6941--6948. IEEE (2019).
  \doi{10.1109/IROS40897.2019.8968116},
  \url{https://doi.org/10.1109/IROS40897.2019.8968116}

\bibitem{Ibraheem}
Ibraheem, N., Hasan, M., Khan, R.Z.: Understanding color models: A review. ARPN
  Journal of Science and Technology  \textbf{2} (01 2012)

\bibitem{Kam2015RVizAT}
Kam, H.R., Lee, S.H., Park, T., Kim, C.H.: Rviz: a toolkit for real domain data
  visualization. Telecommunication Systems  \textbf{60},  337--345 (2015)

\bibitem{kitano1997robocup}
Kitano, H., Asada, M., Kuniyoshi, Y., Noda, I., Osawa, E., Matsubara, H.:
  Robocup: A challenge problem for ai. AI magazine  \textbf{18}(1),  73--73
  (1997)

\bibitem{simost}
Lima~Azevedo, C., Oh, S., Marczuk, K., Soh, H., Basak, K., Toledo, T., Peh,
  L.S., Ben-Akiva, M., Deshmukh, N.: Simmobility short-term: An integrated
  microscopic mobility simulator (01 2017)

\bibitem{Luo}
Luo, J., Hubaux, J.P.: A survey of inter-vehicle communication  (2004),
  \url{http://infoscience.epfl.ch/record/28039}

\bibitem{sumo}
Álvarez López, P., Behrisch, M., Bieker-Walz, L., Erdmann, J., Flötteröd,
  Y.P., Hilbrich, R., Lücken, L., Rummel, J., Wagner, P., Wießner, E.:
  Microscopic traffic simulation using sumo (11 2018).
  \doi{10.1109/ITSC.2018.8569938}

\bibitem{intsim}
Maria, A.: Introduction to modeling and simulation. In: Proceedings of the 29th
  Conference on Winter Simulation. p. 7–13. WSC '97, IEEE Computer Society,
  USA (1997). \doi{10.1145/268437.268440},
  \url{https://doi.org/10.1145/268437.268440}

\bibitem{Nimalsiri}
Nimalsiri, N.I., Mediwaththe, C.P., Ratnam, E.L., Shaw, M., Smith, D.B.,
  Halgamuge, S.K.: A survey of algorithms for distributed charging control of
  electric vehicles in smart grid. IEEE Transactions on Intelligent
  Transportation Systems  \textbf{21}(11),  4497--4515 (2020).
  \doi{10.1109/TITS.2019.2943620}

\bibitem{yuv}
Podpora, M., Korba´s, G., Kawala-Janik, A.: Yuv vs rgb – choosing a color
  space for human-machine interaction. Annals of Computer Science and
  Information Systems  \textbf{Vol. 3} (09 2014). \doi{10.15439/2014F206}

\bibitem{ROSPaper}
Quigley, M., Conley, K., Gerkey, B., Faust, J., Foote, T., Leibs, J., Wheeler,
  R., Ng, A.: Ros: an open-source robot operating system. vol.~3 (01 2009)

\bibitem{raj2020survey}
Raj, T., Hashim, F.H., Huddin, A.B., Ibrahim, M.F., Hussain, A.: A survey on
  lidar scanning mechanisms. Electronics  \textbf{9}(5), ~741 (2020)

\bibitem{sadeghnejad2019validation}
Sadeghnejad, S., Khadivar, F., Abdollahi, E., Moradi, H., Farahmand, F.,
  Sadr~Hosseini, S.M., Vossoughi, G.: A validation study of a virtual-based
  haptic system for endoscopic sinus surgery training. The International
  Journal of Medical Robotics and Computer Assisted Surgery  \textbf{15}(6),
  e2039 (2019)

\bibitem{airsim2017fsr}
Shah, S., Dey, D., Lovett, C., Kapoor, A.: Airsim: High-fidelity visual and
  physical simulation for autonomous vehicles. In: Field and Service Robotics
  (2017), \url{https://arxiv.org/abs/1705.05065}

\bibitem{yang}
Yang, Y., Yuhua, P., Zhaoguang, L.: A fast algorithm for ycbcr to rgb
  conversion. IEEE Transactions on Consumer Electronics  \textbf{53}(4),
  1490--1493 (2007). \doi{10.1109/TCE.2007.4429242}

\end{thebibliography}
\
\end{document}